\documentclass[runningheads]{llncs}
\usepackage[T1]{fontenc}
\usepackage{makecell}
\usepackage[numbers]{natbib}
\usepackage{graphicx}
\usepackage{subcaption} 
\usepackage{pgfplots}
\usepackage{amsmath}
\usepackage{natbib}
\usepackage{hyperref}
\usepackage{float}

\let\llncssubparagraph\subparagraph
\let\subparagraph\paragraph
\usepackage[compact]{titlesec}
\let\subparagraph\llncssubparagraph

\titlespacing{\subsubsection}{0pt}{6pt plus 4pt minus 2pt}{3pt plus 1pt minus 0pt}

\usepackage{color}

\usepackage{caption} 
\begin{document}
\title{A Comparison of Deep Learning Methods for Cell Detection in Digital Cytology}
%
%
\author{Marco Acerbis \and 
Nataša Sladoje \and
Joakim Lindblad
}
\authorrunning{M. Acerbis et al.}
\titlerunning{  }
%
\institute{Center for Image Analysis, Dept. of Information Technology,\\ Uppsala University, Sweden \\}

%
\maketitle              
\begin{abstract}
Accurate and efficient cell detection is crucial in many biomedical image analysis tasks. We evaluate the performance of several Deep Learning (DL) methods for cell detection in Papanicolaou-stained cytological Whole Slide Images (WSIs), focusing on accuracy of predictions and computational efficiency. We examine recent \textit{off-the-shelf} algorithms as well as custom-designed detectors, applying them to two datasets: the CNSeg Dataset and the Oral Cancer (OC) Dataset. Our comparison includes well-established segmentation methods such as StarDist, Cellpose, and the Segment Anything Model 2 (SAM2), alongside centroid-based Fully Convolutional Regression Network (FCRN) approaches. We introduce a suitable evaluation metric to assess the accuracy of predictions based on the distance from ground truth positions. We also explore the impact of dataset size and data augmentation techniques on model performance. Results show that centroid-based methods, particularly the Improved Fully Convolutional Regression Network (IFCRN) method, outperform segmentation-based methods in terms of both detection accuracy and computational efficiency. This study highlights the potential of centroid-based detectors as a preferred option for cell detection in resource-limited environments, offering faster processing times and lower GPU memory usage without compromising accuracy.

\end{abstract}

\section{Introduction}
\begin{figure}[t]
    \begin{subfigure}[t]{0.24\textwidth}
        \centering
        \includegraphics[width=\textwidth]{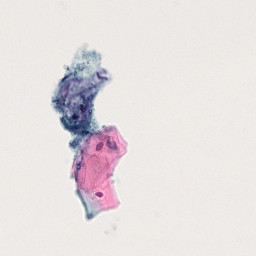}
        \caption{}
    \end{subfigure}
    \begin{subfigure}[t]{0.24\textwidth}
        \centering
        \includegraphics[width=\textwidth]{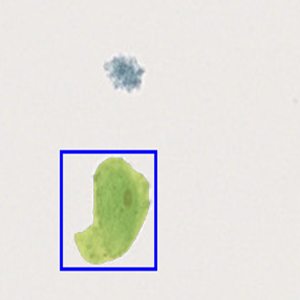}
        \caption{}
    \end{subfigure}
    \centering
    \begin{subfigure}[t]{0.24\textwidth}
        \centering
        \includegraphics[width=\textwidth]{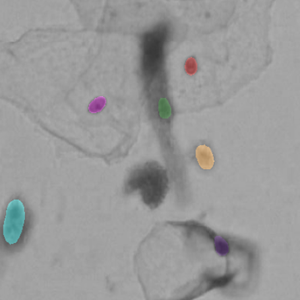}
        \caption{}
    \end{subfigure}
    \begin{subfigure}[t]{0.24\textwidth}
        \centering
        \includegraphics[width=\textwidth]{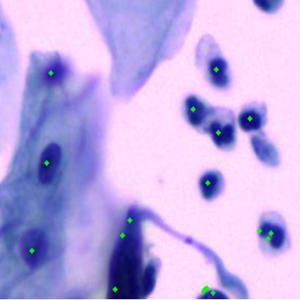}
        \caption{}
    \end{subfigure}
    \caption{(a) Part of cytology image extracted from the OC Dataset; (b) MedSam \citep{ref_medsam} used to segment the object inside a given bounding box; (c) Example of nuclei segmentation using StarDist \citep{ref_StarDist};(d) Cell detection via a centroid-based method (FCRN \citep{ref_Lu_model, ref_FCRN}).}
    \label{fig:techinques_dataset}
\end{figure}

Early stage cancer detection is crucial in order to limit and effectively treat tumor formation. Cytopathological analysis can be a powerful and minimally invasive tool to identify anomalies and suspicious cells. However, to accurately analyze hundreds of thousands of cells in Whole Slide Images (WSIs) is a tedious and difficult task even for the most capable cytologist. DL-based methods allow to identify, extract and classify cells in WSIs, and to support the pathologist in detecting malignancy. The first step in such a pipeline is to detect cells and distinguish them from other formations or external materials. In this work we compare different methods for cell detection, ranging from \textit{off-the-shelf} algorithms to custom designed and trained detectors. An important aspect for cell detection algorithms is the output format that determines how a detected cell is localized. Segmentation based techniques \citep{ref_sam,ref_sam2,ref_StarDist,ref_cellpose, ref_medsam}, as exemplified in Figure \ref{fig:techinques_dataset}(b)-(c), are 
commonly used, and provide pixel-level masks of detected objects of interest. However,
such methods require precise and detailed annotations to be effectively trained. Generating such annotations is time consuming and typically requires the supervision of experts. In many situations, a detailed and precise delineation is not necessary and localizing each cell is enough to analyze it. An often used approach to localize an object in an image is to draw its bounding box \citep{BB_SRPN, ref_BB, ref_BB_yeast, ref_BB_ISE-YOLO}. An even 
less laborious alternative is provided by centroid-based methods. An example is shown in Figure \ref{fig:techinques_dataset}(d). These methods rely only on center-point annotations which are very fast and easy to collect. An example of such a method is proposed in \citet{ref_Lu_model}. We evaluate centroid-based approaches in comparison with alternatives relying on segmentation.

We conduct a series of experiments on two cytology datasets. To evaluate the performances, we develop a custom metric, \textit{Localization Error}, that incorporates the distance between prediction and ground truth in the performance score. In addition, we study the impact of limited training data on performance of non-pretrained methods. 
To facilitate reproducibility, we share our complete implementation and evaluation framework as open source: \url{https://github.com/MIDA-group/Cell-Detection}.

\section{Background and related work}\label{related_work}
Cell detection is an important step in medical image analysis. Development of efficient and fast ways to locate cell nuclei has been stimulated by competitions like the \textit{2018 Data Science Bowl} \citep{2018_Data_Science_Bowl} or the \textit{CoNIC Challenge} \citep{conic_challenge}. The existing approaches include methods that result in (i) cell segmentation, (ii) a bounding box around the cell, and (iii) a centroid of a cell. We briefly describe methods for segmentation and centroid-based detection. We do not include any of the bounding-box detectors, due to their observed weaker performance compared to the other approaches, as presented in \citet{yolo_bad_performance}.

\subsection{Segmentation methods}
One way to perform cell detection involves segmenting the cytoplasm or the nucleus, a technique that provides the most precise and informative detection, but requires costly annotations from experts to accurately train deep learning models. Segmentation of cell (or cell nucleus) enables to derive both its bounding box and its centroid.

\subsubsection{StarDist.} StarDist \citep{ref_StarDist} makes use of star-convex polygons and leverages the capabilities of the U-Net architecture \citep{ref_unet} to rely on the nucleus position rather than the cell boundaries. The proposed method particularly focuses on the challenge of overlapping cells.

\subsubsection{Cellpose.} Proposed in 2020, Cellpose \citep{ref_cellpose} replaces classical methods based on the watershed algorithm for cell segmentation by implementing a U-Net 
architecture. It leverages a set-up for two-channel images: the first common channel is used for the cytoplasmic label, and the second optional channel shares information about the nucleus structure and position. From these inputs, Cellpose generates maps to recognize which pixels should be grouped together in the final masks.

\subsubsection{HoVerNet.}  HoVerNet \citep{hovernet} implements a one-network architecture that simultaneously segments and classifies cells. HoVerNet does not rely on the U-Net architecture, but utilizes its own custom architecture inspired by the Preact-ResNet50 \citep{Preact-ResNet50} model, to generate horizontal and vertical gradient maps used to reconstruct each cell mask.

\subsubsection{MedSAM.} Segment Anything Model (SAM) \citep{ref_sam} affirmed itself among the \textit{state-of-the-art} methods for general image segmentation. \citet{ref_medsam} proposes a modified version, MedSAM, that is fine-tuned on images from different medical fields. MedSAM requires user-drawn bounding box prompts, making it unsuitable for extracting nuclei from batches of images. An example of a segmented cell within a user-defined bounding-box is shown in Figure \ref{fig:techinques_dataset}(b).

\subsubsection{Segment Anything Model 2 (SAM 2).} SAM 2 \citep{ref_sam2} is the second generation of the foundation model Segment Anything and introduces an enhanced architecture to handle both image and video segmentation. Notably, it implements an MAE \citep{ref_MAE} pre-trained Hiera \citep{ref_hiera} to capture high-resolution details, and a Memory Mechanism to handle information from previous and promped frames.

\subsection{Centroids-based methods}
The main drawback of segmentation models is their need for precisely annotated data to be properly trained. However, detailed segmentation is often not required; detected centroids of cell nuclei are often sufficient to locate and cut-out the cells for further processing. Collecting simpler annotations could significantly reduce the required workload and even the need for experts.

\subsubsection{Fully Convolutional Regression Networks (FCRNs).} \citet{ref_Lu_model} proposed to extract the centroids of cell nuclei by means of a Fully Convolutional Regression Network (FCRN) \citep{ref_FCRN} approach. In their pipeline, a modified U-Net \citep{ref_unet} was trained to generate density maps that highlights the position of each detected nucleus. The centroid positions are then extracted from the predicted blobs.
\citet{LetItShine} further develop the FCRN model, reducing the U-Net size and replacing thresholding by local maxima detection, however without presenting any explicit performance evaluation of the performed modifications.

\subsubsection{ACFormer.} In \citet{acformer} the authors underline that cell segmentation is (in many cases) not a necessary step in a cell analysis pipeline. They propose a centroid-based detector that leverages the capabilities of the transformer architecture. The introduced \textit{Affine Consistent Transformer} (ACFormer) aims to locate and classify cell nuclei by leveraging the capabilities of two sub-networks, a local and a global network. The first learns how to handle objects at smaller scales, while the second one handles the large-scale predictions.

\subsubsection{Cell-DETR.} Segmentation algorithms usually struggle to efficiently process large WSIs, especially in resource-limited settings. Cell-DETR \citep{celldetr} proposes to adapt detection transformers (DETR) \citep{detr} as a cost-effective and fast solution to locate and classify cells without the need of segmentation. It uses a hierarchical backbone to generate a four level feature pyramid of the input, that is further processed by a multi-scale DETR \citep{multi_scale_detr} composed of 6 encoder and 6 decoder layers. 

\section{Data}\label{sec:data}
We evaluate the considered methods on two datasets: the \textit{CNSeg Dataset} and the \textit{Oral Cancer (OC) Dataset}. We create four different training/validation/test splits of each dataset to perform a 4-folded cross-validation. 
The validation set is used to tune hyperparameters during the training phase and to calibrate specific algorithms, for example by finding the value of the \textit{radius} parameter for Cellpose. The training set is used to train the centroid-based detectors.

\begin{table}[b]
    \caption{(a) Number of images in train/validation/test sets for each of the four splits of the CNSeg Dataset. (b) Number of ground truth nuclei in each of the four non-overlapping test sets (each containing 477 images) of the CNSeg Dataset. (c) Number of images in train/validation/test sets of the OC Dataset. (d) Number of ground truth nuclei in each test set of the OC Dataset (each containing 78 patches from one out of four WSIs).}
    \begin{subtable}[t]{0.23\textwidth}
        \centering
        \begin{tabular}{c|c}
          Subset  & Images \\
          \hline
          Training & 2462 \\
          Validation & 500 \\
          Test & 477 \\
          \hline
          Total & 3439 \\
        \end{tabular}
         \caption{CNSeg Dataset}
    \end{subtable}
    \begin{subtable}[t]{0.23\textwidth}
        \centering
    \begin{tabular}{c|c}
          Fold & Nuclei \\
          \hline
         1 & 8221\\
         2 & 9075\\
         3 & 7600 \\
         4 & 9419
        \end{tabular}
    \caption{CNSeg Dataset}
    \end{subtable}
    \qquad
    \begin{subtable}[t]{0.23\textwidth}
        \centering
    \begin{tabular}{c|c}
      Subset  & Images \\
      \hline
      Training & 156 \\
      Validation & 78 \\
      Test & 78 \\
      \hline
      Total & 312 \\
    \end{tabular}
        \caption{OC Dataset}
    \end{subtable}
    \begin{subtable}[t]{0.23\textwidth}
        \centering
\begin{tabular}{c|c}
      Fold & Nuclei \\
      \hline
     1 & 381\\
     2 & 364\\
     3 & 571 \\
     4 & 236
    \end{tabular}
        \caption{OC Dataset}
    \end{subtable}
    \label{tab:datasets_summary}
\end{table}

\subsubsection{CNSeg Dataset.}
The CNSeg Dataset \citep{ref_cnseg} is a collections of cervical cell images. All images in this dataset were prepared in a standardized manner: Papanicolaou staining was used, and the images have a resolution of 0.25 $\mu m/px$. Figure \ref{fig:techinques_dataset}(c)-(d) shows segmentation and centroid detection on two images 
from the dataset. 
We use the subset \textit{PatchSeg}, comprising a total of 3439 annotated images of size 512$\times$512 $px$. Tables \ref{tab:datasets_summary}(a)-(b) summarize how we divide the samples in each subset and the total number of ground truth nuclei 
in each test set. Each nucleus annotation consists of a polygon;
we calculated the ground truth centroid position $(C_{x},C_{y})$ as the geometric centroid of the polygon. 


\subsubsection{Oral Cancer Dataset.}
The Oral Cancer (OC) Dataset consists of 312 image patches of size 256$\times$256 $px $ extracted from 4 WSIs obtained from LBC-prepared Papanicolaou stained slides of brush-sampled cells from the oral cavity. The slides have been imaged under white light using a NanoZoomer S60 digital slide scanner, providing a resolution of 0.23 $\mu m/px$. Ground truth centroids in each sample have been annotated by non-specialists using the CytoBrowser \citep{CytoBrowser} tool. Figure \ref{fig:techinques_dataset}(a)-(b) shows examples of tiles of the OC dataset. Tables \ref{tab:datasets_summary}(c)-(d) summarize how the dataset is used to create different subsets for each split and the number of ground truth nuclei in the test sets. Cells from one WSI do not appear in more than one set per fold. 

\begin{figure}[t]
    \centering
    \scalebox{0.70}{
    \begin{tikzpicture}
        \begin{axis}[
            axis lines = middle,
            xlabel = distance,
            ylabel = localization error,
            xmin = 0, xmax = 10,
            ymin = 0, ymax = 2,
            domain=0:10,
            samples=500,
            legend style={at={(1.05,1)}, anchor=north west},
        ]
    
        \addplot[thick, color=green, domain=0:1] {0};
        \addlegendentry{perfect match};
    
        \addplot[thick, color=blue, domain=1:5] {(x-1)/4};
        \addlegendentry{detection};
    
        \addplot[thick, color=red, domain=5:6.2] {(x-1)/4};
        \addlegendentry{missed detection};
    
        \addplot[thick, color=red, domain=6.2:10] {1.3};
    
        \addplot[only marks, mark=*, color=black] coordinates {(1,0)};
        \addplot[only marks, mark=*, color=black] coordinates {(5,1)};
        \addplot[only marks, mark=*, color=green] coordinates {(1,0)};
        \addplot[only marks, mark=*, color=blue] coordinates {(5,0)} ;
        \addplot[only marks, mark=*, color=red] coordinates {(0,1.3)};
        \end{axis}
        
    \end{tikzpicture}}
    \caption{Localization Error $\varepsilon_l$ as a function of distance between detection and ground truth, for $s=1$, $t=5$ and $\alpha = 0.3$.}
    \label{fig:localization_error}
    \end{figure}
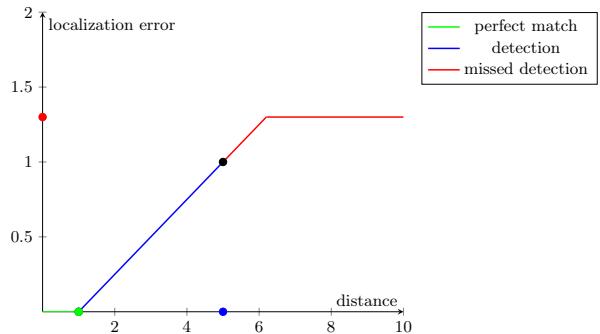

\section{Evaluation metrics}
\subsubsection{Localization Error ($\varepsilon_l$).} A standard follow-up to nucleus detection involves cutting out the cell image using a fixed square window centered on the centroid position. Inspired by previous works (\citet{acformer,metrics}) on centroid-based detection, we define an evaluation metric that accounts for the distance between the predicted centroid and the corresponding ground truth, in addition to measuring missed detections, False Negatives (FN), and extra predictions, False Positives (FP). After calculating the Euclidean distance between each predicted centroid and ground truth centroid, matching is performed using the \textit{Hungarian Algorithm} \citep{ref_hungarian_alg}. Equation \eqref{eq:localization_error} defines how the error is assigned based on the distance $d$ between the ground truth and the matched centroid:
\begin{equation}
\varepsilon_l (d) = max(0, min(1+\alpha, \frac{d-s}{t-s} ) ) \,.
    \label{eq:localization_error}
\end{equation}
\begin{itemize} 
\item If the distance $d$ is less than or equal to the \textit{slack} $s$, defined as 0.25 of the average nucleus diameter, the error is 0, indicating a perfect match;
\item If $d$ is greater than $s$ but less than or equal to the \textit{threshold} $t$, defined as the average nucleus diameter, the error increases linearly with $d$ up to 1;
\item If the distance $d$ exceeds $t$, the detection is considered a miss and both FP and FN counts are incremented by 1. To avoid a discontinuity in the error measure, $\varepsilon_l(d)$ continues to increase linearly until it saturates at $1 + \alpha$. 
\end{itemize}
Figure \ref{fig:localization_error} shows an example of the localization error as a function of the distance from the matched ground truth.

FNs (missed cells) are assigned a Localization Error of 1, whereas
unmatched FPs (spurious detections) are assigned an error of $\alpha$. By varying the value of $\alpha$, we can modulate the cost imposed by FP detections. While FPs are typically handled well by a following classification step, an excessive number can become challenging and computationally too expensive to process. The overall Localization Error $\mathcal{E}_l$ is the sum of Localization Errors for all images divided by the total number $N$ of ground truth nuclei, according to Equation \eqref{eq:total_localization_error}:

\begin{equation}
    \mathcal{E}_l = \frac{1}{N}\sum_N\varepsilon_l \,.
    \label{eq:total_localization_error}
\end{equation}

\begin{figure}[t]
\centering
\includegraphics[width=0.5\textwidth]{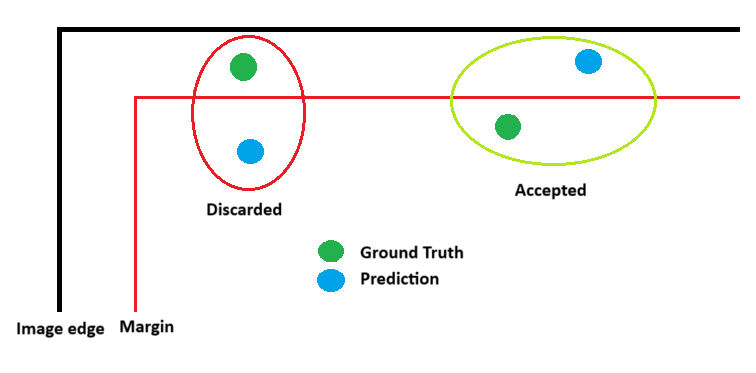}
\caption{Examples of discarded (red, left) and accepted (green, right) detections. If the ground truth lies in the margin, the centroid and any associated prediction are discarded and not included in the analysis.} \label{fig:near_edge}
\end{figure}

\subsubsection{Near-edge detections.} Cells close to the edge of the image may be captured only partially, thus being unsuitable for further analysis. We therefore discard each ground truth nucleus whose distance from the image edge is less than average nucleus diameter. Figure \ref{fig:near_edge} shows that independently from the predicted centroid position, the pair is accepted and included in the results only if the ground truth lies inside the margin. 

\subsubsection{Precision, Recall, and F-Score.} Common metrics to evaluate object detection methods are Precision, Recall and F-score. To make our findings comparable with past and future similar works, we also include these metrics in our analysis.

\subsubsection{Inference rate and GPU memory.} In cytology studies, the usual approach is to process thousands of images in order to extract tens to hundreds of cells from each. To process these large collections of images, machines with dedicated GPUs are typically used. Since the access to such resources is generally limited, we also include the inference rate (IR) and GPU memory usage at inference time in our analysis. 

\section{Methods}
In this section, we describe in more detail how we implement 
the methods introduced in Section \ref{related_work} that we include in our study. 

\subsubsection{Fully Convolutional Regression Network (FCRN).} 
The FCRN approach described in \citet{ref_Lu_model} presents a way to perform nuclei detection relying only on centroid annotations to train a modified U-Net \citep{ref_unet} for which the final softmax is replaced by a linear activation function. From the annotations, ground truth binary masks are constructed and then dilated by a disk of radius $r$, followed by convolution with a 2D Gaussian filter to generate a fuzzy ground truth $D$. The FCRN learns the mapping between the original image and the corresponding fuzzy ground truth. At inference, the model predicts a probability map for each image which is binarized at a threshold $T$ and the centroid of each connected foreground component is extracted as a detected nucleus location.

In our tests, we replicate the U-Net architecture of \citep{ref_Lu_model}\footnote{\url{https://github.com/MIDA-group/OralScreen}}, converting the original implementation to the most recent \textit{Keras} release to overcome compatibility issues. We also implement the centroid extraction using Python instead of ImageJ Macro. We train the model following the authors instructions by generating ground truth binary masks around the centroid position, followed by Gaussian blurring. Each training run lasts 100 epochs and has a batch size of 32. A relevant hyperparameter for this model is the \textit{threshold} value $T$ used to binarize the network output. Based on empirical evaluation on the validation set, we set it to 0.58 for images from the CNSeg dataset, and to 0.65 for the OC dataset.

\subsubsection{Improved FCRN (IFCRN).}\label{our_method} 
It is observed that the binarization of the network output with a global threshold $T$ does not provide reliable results for heterogeneous dataset. In \citet{LetItShine}, nuclei locations are instead detected at local maxima of height $h>0.5$ in the prediction output, leading to improved performance. Further, the original 23 convolution layers of the U-Net are reduced to 8 convolution layers, providing a much leaner model. Images are downsampled by a factor 4$\times$4, further reducing the computational burden. The two steps for generating the fuzzy ground truth of \citet{ref_Lu_model}, dilation followed by Gaussian blur, are replaced by only Gaussian blur ($\sigma=3$). 

In our tests, we replicate the architecture of \citep{LetItShine}.
Different from \citep{LetItShine}, we work with only one focus level, to reproduce the results on images from different datasets, like the CNSeg Dataset \citep{ref_cnseg}. Further, we tune the detection sensitivity by adjusting the required minimal height $h$ for the detected local maxima. Based on empirical evaluation on the validation set, we set it to 0.4 for images from both the CNSeg and the OC datasets.

\subsubsection{Cellpose.} Cellpose \citep{ref_cellpose} is made available through a Python library with the same name that relies on PyTorch. There are a two different models: \textit{"cyto"} and \textit{"nuclei"}. 
The more complex \textit{"cyto"} model, which uses both cell boundaries and nuclei information to segment the cell, performed significantly better on our validation sets, and we therefore use that model for our evaluation.
The value of the parameter \textit{diameter} requested by the algorithm is set to 30 on the CNSeg dataset and to \textit{Auto} on the OC dataset. In this second case, the helper functions of Cellpose automatically calculate the value of the the parameter. We select these settings since they obtained the best performance on the validation set.

\subsubsection{StarDist.} StarDist \citep{ref_StarDist} is released in two different versions, accessible via its own Python library, and it is implemented with Tensorflow. One is trained on brightfield data of H\&E stained cells extracted from the \textit{MoNuSeg} 2018 \citep{MoNuSeg2018} dataset and the \textit{TNBC} dataset presented in \citet{TNBC}. The second model is trained on fluorescence images using nuclear markers extracted from the \textit{2018 Data Science Bowl} \citep{2018_Data_Science_Bowl} dataset. Since both datasets in our study collect brightfield images, we select the former model for our experiments.

\subsubsection{Segment Anything Model 2.} Segment Anything 2 \citep{ref_sam2} model and weights can be directly downloaded from the official FAIR GitHub repository\footnote{\url{https://github.com/facebookresearch/sam2}}. It leverages a transformer architecture trained on the SA-V dataset \citep{ref_sam2}. We study SAM2 L and SAM2 T. We include both versions because \citet{ref_sam2_med_img} highlights the superior overall performance of the non-fine-tuned large model compared to the tiny counterpart, whereas \citet{ref_medsam} focuses on a fine-tuned SAM Tiny \citep{ref_sam}.

\section{Experiments and results}
In this section we present the results of the evaluation experiments. All experiments are performed on an Intel(R) Core(TM) i9-9940X running Gentoo Linux 6.6.67 with  Python 3.12.8, PyTorch 2.5.1+cu124, and Tensorflow 2.18.0. The machine is equipped with a Nvidia GeForce RTX 4090 GPU card with 24 GB of memory. 

\begin{figure}[t]
    \begin{subfigure}[t]{0.5\textwidth}
        \centering
        \includegraphics[width=\linewidth]{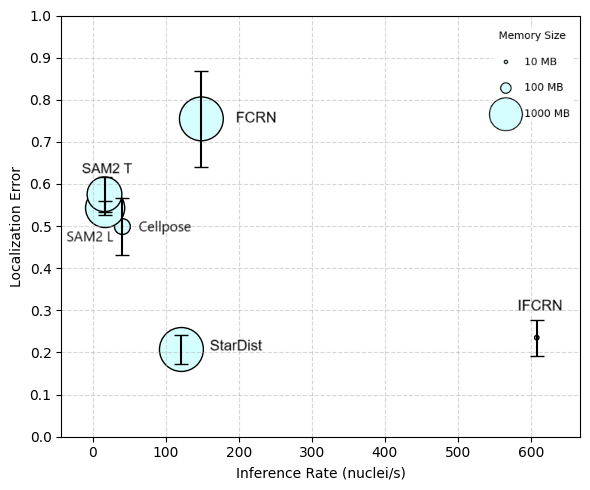}
        \caption{CNSeg Dataset}
    \end{subfigure}
    \begin{subfigure}[t]{0.5\textwidth}
        \centering
        \includegraphics[width=\linewidth]{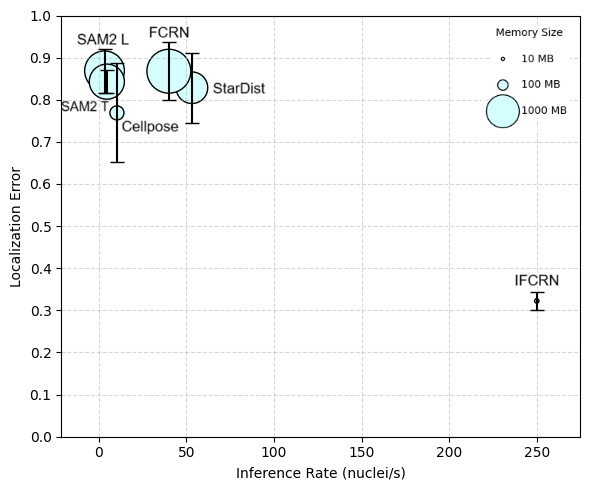}
        \caption{OC Dataset}
    \end{subfigure}
    \caption{Localization Error $\mathcal{E}_l$ ($\alpha=0.3)$ for the evaluated methods on (a) the CNSeg Dataset and on (b) the OC Dataset. Best performance is located in the right bottom part of the graph. The size of the circle for each method is proportional to the memory requirements of that model.}  
    \label{fig:results}
\end{figure}

\subsection{Comparison of different detection techniques}\label{results_exp1}
The main experiments conducted in this study aim to compare the performance of different methods on the task of cell detection in cytology images. For both datasets, CNSeg and OC, we implemented the following pipeline:
\begin{itemize}
    \item \textbf{Parameter tuning.} When needed, we select the values for a given method parameters by running different configurations on the validation set. The best performing configuration are then selected to run the experiments;
    \item \textbf{Zero-shot evaluation.} Segmentation methods are evaluated in zero-shot experiments without any re-training or fine-tuning;
    \item \textbf{Training from scratch.} For each split, FCRN and IFCRN are \textit{trained from-scratch} and then evaluated on the corresponding test set.
\end{itemize}
The results, reported in Tables \ref{tab:cnseg_results} and \ref{tab:OC_results}, are the weighted average of the results for each split, where the number of ground truth nuclei in the test set, reported in Table \ref{tab:datasets_summary}(b) and (d), is used as weight. Each result is followed by $\pm$ the standard deviation computed over the four folds. 

\subsubsection{CNSeg results.} From Table \ref{tab:cnseg_results} and Figure \ref{fig:results}(a) we observe that the overall best performing methods on the CNSegn Dataset are StarDist and IFCRN. IFCRN achieves comparable results to StarDist and the overall best F-Score. It is also much faster with an inference rate of 608 nuclei/$s$, and requires only a fraction of the GPU memory at inference time of the other methods. 

\begin{table}[t]
\centering
\caption{Results on the CNSeg dataset. Inference rate (IR) is measured in nuclei/$s$. Memory refers to the GPU memory required at inference time.}
\begin{tabular}{|c|c|c|c|c|c|c|c|}
\hline
\textbf{Model} & 
\textbf{$\mathcal{E}_l$ ($\alpha=0.3$)} & 
\textbf{$\mathcal{E}_l$ ($\alpha=1$)} &
\textbf{Precision} & 
\textbf{Recall} & 
\textbf{F-Score} & 
\makecell{\textbf{IR} \\ \textbf{(n/s)}} & 
\makecell{\textbf{Memory} \\ \textbf{(MB)}} \\ 
\hline
Cellpose & 0.50 $\pm$0.07 & 0.67 $\pm$0.09 & 79.3 $\pm$3.0 & 76.5 $\pm$2.9 & 77.8 $\pm$2.8 & 41 & 526\\
\hline
StarDist & \textbf{0.21} $\pm$0.03 & \textbf{0.31} $\pm$0.06 & 86.2 $\pm$3.4 & \textbf{91.2} $\pm$1.6 & 88.6 $\pm$2.5 & 121 & 4317\\
\hline 
SAM2 L & 0.54 $\pm$0.02 & 0.74 $\pm$0.03 & 77.2 $\pm$2.4 & 75.8 $\pm$0.85 & 76.7 $\pm$1.0 & 17 & 3418 \\
\hline
SAM2 T & 0.57 $\pm$0.04 & 0.75 $\pm$0.05 & 81.1 $\pm$2.5 & 73.0 $\pm$2.4 & 76.8 $\pm$1.0 & 16 & 2687\\
\hline
FCRN  & 0.75 $\pm$0.11 & 0.79 $\pm$0.12 & \textbf{94.7} $\pm$2.4 & 60.6 $\pm$4.3 & 73.8 $\pm$3.7 & 149 & 4297 \\
\hline
IFCRN &  0.23 $\pm$0.04 & 0.32 $\pm$0.05 & 88.8 $\pm$1.8 & 89.1 $\pm$2.6 & \textbf{88.9} $\pm$2.0 & \textbf{608}  & \textbf{3}\\
\hline

\end{tabular}
\label{tab:cnseg_results}
\end{table}

\begin{table}[t]
\centering
\caption{Results on the OC dataset. Inference rate (IR) is measured in nuclei/$s$. Memory refers to the GPU memory required at inference time.}
\begin{tabular}{|c|c|c|c|c|c|c|c|}
\hline
\textbf{Model} & 
\textbf{$\mathcal{E}_l$ ($\alpha=0.3$)} & 
\textbf{$\mathcal{E}_l$ ($\alpha=1$)} &
\textbf{Precision} & 
\textbf{Recall} & 
\textbf{F-Score} & 
\makecell{\textbf{IR} \\ \textbf{(n/s)}} & 
\makecell{\textbf{Memory} \\ \textbf{(MB)}} \\ 
\hline
Cellpose & 0.77 $\pm$0.12 & 1.2 $\pm$0.29 & 64.6 $\pm$11 & 72.1 $\pm$4.5 & 67.8 $\pm$6.7 & 10 & 403\\
\hline
StarDist & 0.83 $\pm$0.08 & 0.86 $\pm$0.09 & \textbf{93.7} $\pm$3.3 & 60.5 $\pm$5.2 & 73.4 $\pm$4.0 & 53 & 2203\\
\hline 
SAM2 L & 0.87 $\pm$0.05 & 1.7 $\pm$0.1 & 47.4 $\pm$1.6 & 71.1 $\pm$2.7 & 56.9 $\pm$1.7 & 3 & 3506 \\
\hline
SAM2 T & 0.84 $\pm$0.03 & 1.3 $\pm$0.1 & 61.4 $\pm$5.6 & 65.7 $\pm$3.2 & 63.3 $\pm$2.0 & 5 & 2726\\
\hline
FCRN  & 0.87 $\pm$0.07 & 0.99 $\pm$0.13 & 86.6$\pm$8.5 & 60.8 $\pm$2.3 & 71.3 $\pm$3.4  & 40 & 4293 \\
\hline
IFCRN &  \textbf{0.32} $\pm$0.02 & \textbf{0.50} $\pm$0.14 & 80.5 $\pm$1.1 & \textbf{86.6} $\pm$2.3 & \textbf{83.0} $\pm$6.4 & \textbf{250}  & \textbf{1}\\
\hline
\end{tabular}
\label{tab:OC_results}
\end{table}

\subsubsection{OC results.} Table \ref{tab:OC_results} and Figure \ref{fig:results}(b) present the results for the OC Dataset. For this dataset IFCRN turns out to be the best performing method. The $\mathcal{E}_l$ for $\alpha=0.3$ is 42\% better than the results for the second best performing model, Cellpose. At the same time, IFCRN inference rate is 5 times the second fastest method, StarDist, while only using a small fraction of GPU memory at inference time.

\subsection{Impact of the amount of training data} \label{results_exp2}
\textit{Off-the-shelf} methods, and foundation models (such as SAM 2) in particular, are trained on large datasets, making them a powerful tool even when fine-tuning is not possible. Smaller customizable models, like FCRN, can be trained on relatively small annotated data. In this experiment we use training sets, extracted from the CNSeg Dataset \citep{ref_cnseg}, of varying size to train IFCRN. Each training run has 450 epochs and a batch size of 32.

\subsubsection{Data augmentation.} A key aspect to achieve good results is the amount of data available. It is often beneficial to include data augmentation to train a more robust model. We explore different augmentations; from simple \textit{Random Rotations} and \textit{Flips} to \textit{Color Jitter}, \textit{GaussianBlur}, and \textit{GaussianNoise}.

\begin{figure}
    \centering
    \includegraphics[width=0.8\linewidth]{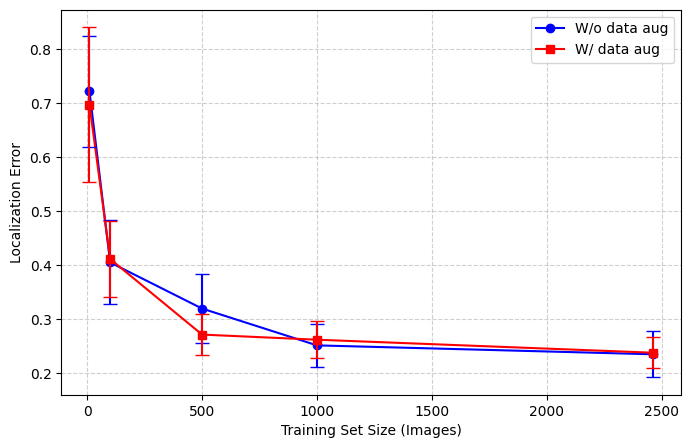}
    \caption{Impact of different size of the training set for IFCRN on the  CNSeg Dataset, without and with data augmentation.}
    \label{fig:training_set_size}
\end{figure}

\subsubsection{Results.} Figure \ref{fig:training_set_size} shows the positive impact of increasing the number of samples in the training set. IFCRN results are comparable to those of Cellpose and SAM2 with only 100 images in the training set, while with 1000 images IFCRN reaches results in the range of the top performing models, like StarDist. As expected, data augmentations can contribute to improve the results, especially when the number of samples is very limited.

\begin{figure}[t]
    \begin{subfigure}[t]{0.192\textwidth}
        \centering
        \includegraphics[width=0.98\textwidth]{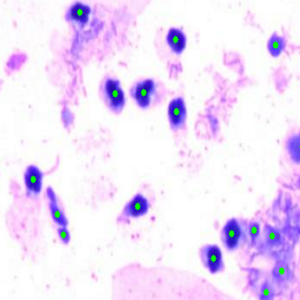}
        \caption{}
    \end{subfigure}
    \begin{subfigure}[t]{0.192\textwidth}
        \centering
        \includegraphics[width=0.98\textwidth]{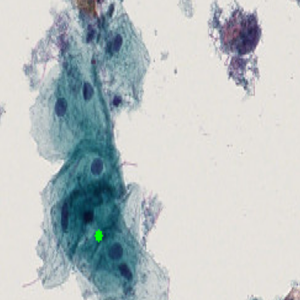}
        \caption{}
    \end{subfigure}
    \centering
    \begin{subfigure}[t]{0.192\textwidth}
        \centering
        \includegraphics[width=0.98\textwidth]{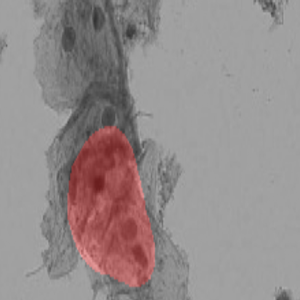}
        \caption{}
    \end{subfigure}
        \begin{subfigure}[t]{0.192\textwidth}
        \centering
        \includegraphics[width=0.98\textwidth]{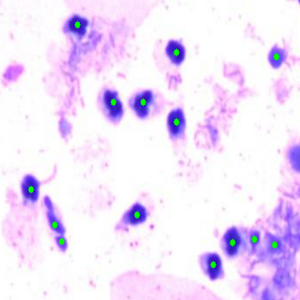}
        \caption{}
    \end{subfigure} 
        \begin{subfigure}[t]{0.192\textwidth}
        \centering
        \includegraphics[width=0.98\textwidth]{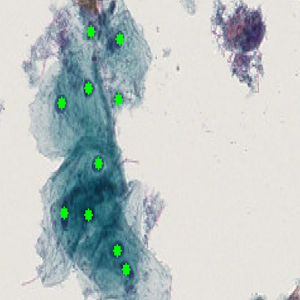}
        \caption{}
    \end{subfigure}
    \caption{Example results of StarDist and IFCRN on images from CNSeg and OC Datasets: (a) StarDist successfully detects the nuclei in an image from the CNSeg Dataset. (b) StarDist fails to locate the cell nuclei in an image from the OC Dataset. (c) Failed segmentation attempt of StarDist on the same OC image. (d) IFCRN successfully detects the nuclei in an image from the CNSeg Dataset. (e) IFCRN successfully detects the nuclei in an image from the OC Dataset.}
    \label{fig:success_fails}
\end{figure}

\section{Discussion}
From the results presented in Section \ref{results_exp1}, we observe a much reduced performance of all the segmentation methods on OC data, compared to CNSeg Dataset. Figure \ref{fig:success_fails} shows two examples. Images extracted from the CNSeg Dataset generally present well defined and isolated cells whose nuclei can also be easily identified; possibly explaining the good results of the segmentation algorithms such as StarDist, as shown in Figure \ref{fig:success_fails}(a). On the contrary, images from the OC Dataset may present groups of overlapping cells, whose boundaries and nuclei are not always clear and easy to recognize. This may cause segmentation algorithms to fail in properly detecting the cell nucleus as shown in Figure \ref{fig:success_fails}(b)-(c). Both FCRN and IFCRN maintain similar results on both datasets. This result can be explained by how these methods approach the task: instead of outlining the cell boundary, that is not always well defined, they search for the usually more highlighted nucleus. This way they can still perform well even when the cells are clustered or overlapping. Figure \ref{fig:success_fails}(d)-(e) shows two example of IFCRN results on images from both datasets.\\

A WSI typically contains 10,000-150,000 cells, making inference speed important when processing multiple WSIs. In terms of inference rate and GPU memory usage, IFCRN is by far the best performing method. While it is expected that a foundation model such as SAM 2 would be more computationally demanding, even other U-Net based architectures under-perform. For Cellpose or StarDist this may be due to the extra resources and time needed to generate segmentation masks, from which sub-regions the centroids are extracted. FCRN fails to efficiently implement nucleus detection by using a deeper network than IFCRN and by implementing a threshold-based extraction of the predicted centroids. \\

From the second experiment, presented in Section \ref{results_exp2}, we observe that not only, as expected, more samples in the training set leads to improved results, but also that already rather small datasets are enough to obtain results comparable to, or better than those of \textit{off-the-shelf} methods. This important result shows how custom solutions trained on small dataset provide a valid and strong alternative to more generally (pre-)trained solutions. The impact of data augmentations can also be relevant, especially with very limited datasets, even if it requires more testing and parameter tuning to find the optimal configuration.

\section{Conclusion}
We present a comparison between contemporary deep learning-based segmentation and centroid-based cell detectors. The aim of our study is to find efficient solutions to extract cells from WSIs for further analysis. We observe that centroid-based methods, and in particular the IFCRN method, perform on par or better than segmentation-based approaches, especially when cells are clustered or overlapping. The IFCRN method delivers the overall best performance in our tests performed. The IFCRN method also requires much less computational power and can process nuclei up to 50$\times$ faster than the other methods. 

We observe that, under the constraint of limited data, a simpler and more tailored solution may be a better choice for the task of cell detection, reaching results comparable with those of more complex pre-trained method.



\begin{credits}
\subsubsection{\ackname} 
This work is supported by: Sweden’s Innovation Agency (VINNOVA) through the Analytic Imaging Diagnostics Arena (AIDA), grant 2021-01420, the Swedish Research Council, grant 2022-03580, and Cancerfonden projects 22 2353 Pj and 22 2357 Pj. We are grateful to M. Matić for accurate cell location annotations.

\subsubsection{\discintname}
The authors have no competing interests to declare that are relevant to the content of this article.
\end{credits}

{\small
\renewcommand{\bibname}{References}
\bibliographystyle{plainnat}
\bibliography{references}
}
\end{document}